\documentclass[runningheads, envcountsame, a4paper]{llncs}
\usepackage{subfigure}
\usepackage{bm}
\usepackage{algorithm}
\usepackage{amsfonts}
\usepackage[noend]{algpseudocode}
\usepackage{mathtools}
\usepackage[misc]{ifsym}
\usepackage{filecontents}
\usepackage{amssymb}
\usepackage{amsmath}
\usepackage{commath}
\usepackage{dsfont}
\DeclareMathOperator*{\argmax}{arg\,max}

\usepackage[hyphens]{url}
\usepackage[pdftex]{graphicx}
\usepackage{epstopdf}
\begin{document}
\title{Generating Multi-type Temporal Sequences to Mitigate Class-imbalanced Problem}
\titlerunning{Generating Multi-type Temporal Sequences}
%
\author{Lun Jiang\thanks{Authors contributed equally.}\and
 Nima Salehi Sadghiani \inst{*} (\Letter) \and
 Zhuo Tao \inst{*} \and Andrew Cohen  }

\authorrunning{L. Jiang et al.}
 \institute{Unity, San Francisco CA 94103, USA \\ 
\email{$\{$lun,nimas,zhuo,andrew.cohen$\}$@unity3d.com}}
\toctitle{Generating Multi-type Temporal Sequences to Mitigate Class-imbalanced Problem}
\tocauthor{Lun~Jiang, Nima~Salehi Sadghiani,Zhuo~Tao,Andrew~Cohen}
\maketitle \setcounter{footnote}{0}
%

\begin{abstract}
From the ad network standpoint, a user's activity is a multi-type sequence of temporal events consisting of event types and time intervals. Understanding user patterns in ad networks has received increasing attention from the machine learning community. Particularly, the problems of fraud detection, Conversion Rate (CVR), and Click-Through Rate (CTR) prediction are of interest. However, the class imbalance between major and minor classes in these tasks can bias a machine learning model leading to poor performance. This study proposes using two multi-type (continuous and discrete) training approaches for GANs to deal with the limitations of traditional GANs in passing the gradient updates for discrete tokens. First, we used the Reinforcement Learning (RL)-based training approach and then, an approximation of the multinomial distribution parameterized in terms of the softmax function (Gumble-Softmax). Our extensive experiments based on synthetic data have shown the trained generator can generate sequences with desired properties measured by multiple criteria.
\keywords{Multi-type sequences \and Temporal events \and Generative adversarial network \and Reinforcement learning}
\end{abstract}
\section{Introduction}
Game developers can monetize their games by selling in-game ad placements to advertisers. Ads can be integrated in multiple ways such as a banner in the background or commercials during breaks (when a specific part of the game is completed). There are four main elements in the game advertising ecosystem: publishers or developers, advertisers, advertising networks, and users \cite{mouawi2019crowdsourcing}. Game advertising networks connect advertisers with game developers and serve billions of ads to user devices, triggering enormous ad events. For example, Unity Ads reports 22.9B+ monthly global ad impressions, reaching 2B+ monthly active end-users worldwide  \footnote{\url{https://www.businesswire.com/news/home/20201013005191/en/}}.

An ad event is a user interaction e.g. request, start, view, click, and install. Each type stands for one specific kind of ad-related user action happening at a specific time. A complete ad life cycle consists of a temporal sequence of ad events, each of which is a tuple of event types with corresponding time intervals. Click and install are two kinds of ad events commonly associated with ad revenue. Pay-Per-Click \cite{kapoor2016pay} and Pay-Per-Install \cite{thomas2016investigating} are the most widely used advertising models for pricing.

Unlike traditional advertising, online advertising offers services that link user interactions to conversions or clicks. Due to this, predicting a user's probability of clicking or conversion rate has become one of the most important problems in online advertising \cite{chapelle2014simple}. Predicting Conversion Rate (CVR) and  Click-Through Rate (CTR) are usually treated as supervised learning problems \cite{deng2021calibrating}. For example, in CTR prediction, the labels are click/not-click an ad for every user. The sequence of events before a click/not-click response are used as features of the supervised learning model.

Unfortunately, as advertisers allocate more of their budget into this ecosystem, there is more incentive to abuse the advertising networks and defraud advertisers of their money \cite{nagaraja2019clicktok}. Fraudulent ad activity aimed at generating illegitimate ad revenue or unearned benefits are one of the major threats to online advertising models. Common types of fraudulent activities include fake impressions \cite{haider2018ensemble}, click bots \cite{haddadi2010fighting,kudugunta2018deep}, or click farms \cite{oentaryo2014detecting}. 

Given the massive ad activity data in-game advertising networks, machine learning-based approaches have become popular in the industry. However, it is not a straightforward task to train machine learning models directly on the sequences collected from ad activities \cite{choi2020identifying}. 

The primary issue in these problems is class imbalance. By definition, the ratio of typical user behavior to anomalous will heavily favor typical. For example, the CVR can be as low as $0.01\%$ for game ads. Similarly, most ad traffic is non-fraudulent, and data labeling by human experts is time-consuming. In these scenarios, label sparsity leads to low availability of labeled sequences for the minor class. Simply oversampling the minority class can cause significant overfitting, while undersampling the majority may lead to information loss and yield a tiny training dataset \cite{ba2019improving}. In this study, we present a novel method to generate synthetic data to mitigate class imbalance.

The main contributions of our work can be summarized as follows:
\begin{enumerate}
    \item A novel reinforcement learning formulation that trains a generator to generate multi-type temporal sequences with non-uniform time intervals.
    \item A novel training method for sequence GAN that uses a critic network.
    \item A new application for event-based sequence GAN in game advertising.
\end{enumerate}

\section{Related Work}
Generative Adversarial Networks (GANs) \cite{goodfellow2014generative} have drawn significant attention as a framework for training generative models capable of producing synthetic data with desired structures and properties \cite{killoran2017generating}. It was proposed to use GANs to generate data that mimics training data as an augmented oversampling method with an application in credit card fraud. The generated data is used to assist the classification of credit card fraud \cite{ba2019improving}. 

\subsection{GAN for Sequence Data}
Despite the remarkable success of GANs in generating synthetic data, very few studies focus on generating sequential data. This is due to additional challenges in generating temporally dependent samples. Recurrent Neural Network (RNN) solutions are state-of-the-art in modeling sequential data. Recurrent Conditional GAN (RCGAN) generates real-valued multi-dimensional time series and then uses the generated series for supervised training \cite{esteban2017real}. The time series data in their study were physiological signals sampled at specific fixed frequencies. However, ad event data has higher complexity due to non-uniform time intervals and discrete event types and thus can not be modeled as wave signals. In ad event sequences, two events with a short time interval tend to be more correlated than events with larger time intervals.

A GAN-based generative model for DNA along with an activation maximization technique for DNA sequence data is proposed by \cite{killoran2017generating}. Their experiments have shown that these generative techniques can learn the important structure from DNA sequences and can be used to design new DNA sequences with desired properties. Similarly to the previous study, their focus is on fixed interval sequences. 

The Long Short-Term Memory (LSTM)-Autoencoder is used to encode the benign users into a latent space \cite{zheng2019one}. They proposed using One-Class Adversarial Network (OCAN) for the training process of the GAN model. In their training framework, the discriminator is trained to be a classifier for distinguishing benign users, and the generator produces samples that are complementary to the representations of benign users.

\subsection{RL for GANs with Sequences of Discrete Tokens}
When generating continuous outputs, gradient updates can be passed from the discriminator to the generator. However, for discrete outputs, this is not straightforward due to a lack of differentiability. The issue of training GAN models to generate sequences of discrete tokens is addressed in \cite{yu2017seqgan}. They proposed a sequence generation framework called SeqGAN that models the data generator as a stochastic policy learned via Reinforcement Learning (RL) \cite{sutton2018reinforcement}. SeqGAN learns a policy using the vanilla policy gradient and Monte Carlo (MC) rollouts to approximate the advantage. MC rollouts are a computationally expensive process in the training loop. Moreover, SeqGAN is limited to discrete token generation. In our work, we propose a modified version of SeqGAN that can generate both discrete tokens and continuous time-intervals. Additionally, to efficiently train the policy network, we employ a Critic network to approximate the return given a partially generated sequence to speed up the training process. This approach also brings the potential to use a trained Critic network for early fraud detection from partial sequences. 

An application of SeqGAN in recommendation systems is presented in \cite{zhao2020adversarial}. The paper solves the slow convergence and unstable RL training by using the Actor-Critic algorithm instead of MC roll-outs. Their generator model produces the entire recommended sequences given the interaction history while the discriminator learns to maximize the score of ground-truth and minimize the score of generated sequences. In each step, the generator $G$ generates a token by top-k beam search based on the model distribution. In our work, we directly sample from the distribution of the output probabilities of the tokens. While our methodologies are close, we are aiming for different goals. We optimize the generated data to solve the sample imbalance problem while they optimize for better recommendations. Therefore, different evaluation metrics are needed. Our methodologies also differ in the training strategy. For example, we used a Critic network as the baseline, whereas they used Temporal-Difference bootstrap targets. They pre-trained the discriminator on the generated data to reduce the exposure bias, while we pre-trained the discriminator on the actual training data for improving the metrics we use in our experiments. More importantly, they do not include time intervals as an attribute in their model while we have time intervals in our models.
 
The idea of using SeqGan to adversarially learn the output sequences while optimizing towards chemical metrics with the algorithm REINFORCE \cite{sutton2018reinforcement} is proposed in \cite{guimaraes2017objective}. They have shown that it is often advantageous to guide the generative model towards some desirable characteristics, while ensuring that the samples resemble the initial distribution.

\subsection{Gumbel-Softmax Distribution for GANs with Sequences of Discrete Tokens}
The Gumbel-Softmax distribution is proposed in \cite{kusner2016gans} to address the limitation of GANs for generating sequences of discrete tokens. The Gumbel-Softmax is a continuous approximation to a multinomial distribution parameterized over a softmax function. This approximation is differentiable thus enabling backpropagation through an approximation of a discrete sampling procedure. A temperature parameter can be used to controll the degree of approximation  \cite{jang2016categorical}. When the temperature is lower, the approximation is closer to the one hot distribution; when it is higher, the approximation is closer to a uniform distribution. 

Another application of Gumbel-Softmax distributions is proposed in \cite{de2018molgan} for generating small molecular graphs.

\section{Methodology}
In this section, we introduce a new methodology to generate multi-type sequences using GAN, which can be trained by using RL and Gumbel-Softmax reparametrization.

\subsection{Definitions}\label{def}
The sequence of an ad event with length $L$ is composed of two sub-sequences, the sub-sequence of event types $\bm{x}$ and the sub-sequence of time stamps. First, we transform the time stamps $\bm{t}$ into time intervals $\bm{\Delta \textbf{t}}$ and $\Delta t_{m} = t_{m} - t_{m-1}, \forall m \in [1, L]$, and $\Delta t_1 = t_{1} - 0$. 
Then, we combine the event types and time intervals into a joint multi-type sequence $\bm{A}$: 
\begin{equation*}
    \bm{A} = \bm{A}_{1:L} \nonumber\\
    = \{(x_1,\Delta t_1),(x_2,\Delta t_2),\dots, \Delta (x_m,\Delta t_m), \dots,(x_L,\Delta t_L) \} \label{eq:complete_sequence}
\end{equation*}
\noindent where a bold $\bm{A}_{1:m}$ denotes a partial sequence from step $1$ to step $m$, and a non-bold $A_{m} = (x_m, \Delta t_m)$ denotes a single pair in the sequence.

\subsection{RL and Policy improvement to train GAN} \label{sec:rl_train}
\begin{figure}[tb]
  \centering
  \includegraphics[width=\linewidth]{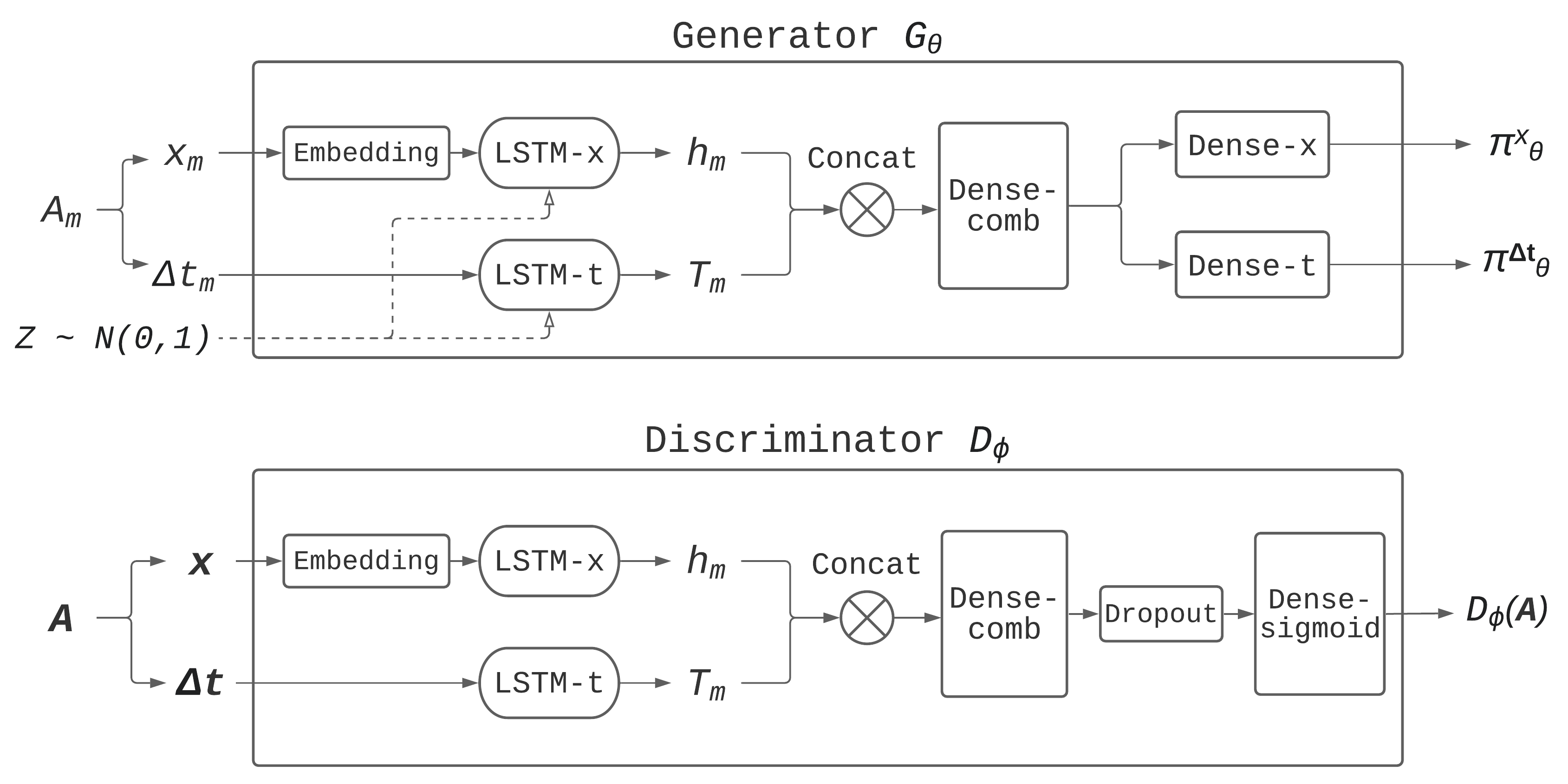}
  \caption{Architecture of the Generator and the Discriminator.}
  \label{fig:model-arch}
\end{figure}

We implemented a modified version of SeqGAN model to generate multi-type temporal sequences. The architecture is shown in Fig. \ref{fig:model-arch}. 

The sequence generation process of our generator $G$ can be modeled as a sequential decision process in RL. $h_m$ and $T_m$ are the hidden states of LSTM cells, and $Z \sim \mathcal{N}(0, 1)$ is the normal noise used to initialize $h_m$ and $T_m$ at the beginning of each generation process. 

From the perspective of RL, at each step $m$, we define the state $S_m$ as the partial sequence $\bm{A}_{1:m}$, a.k.a, \begin{equation}
   S_m = \bm{A}_{1:m} \label{equ:state_def}
\end{equation}
\noindent
During the generation process, at each step $m$, a new pair
\begin{equation}
   {A}_{m+1} = (x_{m+1}, \Delta t_{m+1})
\end{equation}
\noindent is appended to the current partial sequence $\bm{A}_{1:m}$ to formulate a new partial sequence $\bm{A}_{1:m+1}$, and thus transit to a new state $S_{m+1}$, based on the definition of state in (\ref{equ:state_def}). This process repeats step by step, until a complete sequence $\bm{A}$ of length $L$ described in (\ref{eq:complete_sequence}) is fully constructed.

To make decisions in this sequence generation process, we employ a hybrid policy to represent action spaces with both continuous and discrete dimensions (similar to the idea in \cite{neunert2020continuous}). This policy is designed to choose discrete event types and continuous time intervals, assuming their action spaces are independent. Then we use a categorical distribution and a Gaussian distribution to model the policy distributions for the event types and the time intervals respectively. So the hybrid generator policy can be defined as:

\begin{align}
   G_{\theta}(a_m|S_m) \nonumber = & \pi^{x}_{\theta}(a^{x}_m|S_m) \cdot \pi^{\Delta t}_{\theta}(a^{\Delta t}_m|S_m) \\
   = & Cat(x | \alpha_{\theta}(S_{m})) \cdot \mathcal{N}(\Delta t| \mu_{\theta}(S_{m}), \sigma^2_{\theta}(S_{m})) \label{equ:hybrid_policy}
\end{align}

\noindent where $x \in \bm{K}, \Delta t \in R_{\geq 0}$. $\bm{K}$ is the set of all event types.
Then an action $a_m$ is taken at step $m$ to sample the next event type $x_{m+1}$ and the next time interval $\Delta t_{m+1}$ given the hybrid policy (\ref{equ:hybrid_policy}). So the action has discrete part and the continuous part sampled independently: 
\begin{align}
   a_m &= \{a^{x}_m, a^{\Delta t}_m\} \label{equ:action_def} \\
   a^{x}_m &= x_{m+1} \sim Cat(x | \alpha_{\theta}(S_{m}))  \\
   a^{\Delta t}_m &= {\Delta t}_{m+1} \sim \mathcal{N}(\Delta t| \mu_{\theta}(S_{m}), \sigma^2_{\theta}(S_{m})) \label{equ:action_sampling}
\end{align}
where $a^{x}_m$ is the action to find the next event type $x_{m+1}$ and $a^{\Delta t}_m$ is the action to find the next time interval $\Delta t_{m+1}$.

When generating a new event type and time interval at each step, we follow the generator policy and sample from categorical and Gaussian distributions independently and concatenate them to obtain the action vector $a_m$, then append them to the current partial sequence $\bm{A}_{1:m}$ to obtain a new partial sequence $\bm{A}_{1:m+1}$. Once a complete sequence of length $L$ has been generated, we pass the sequence $\bm{A}$ to the Discriminator $D$ which predicts the probability of the sequence to be real against fake:
\begin{equation}
   D_{\phi}(\bm{A}) = Pr(Y = 1 | \bm{A} ; \phi)
\end{equation}

The feedback from $D$ can be used to train $G$ to generate sequences similar to real training data to deceive $D$. Because the discrete data is not differentiable, gradients can not passed back to generator like in image-base GANs.
 
The original SeqGAN training uses Policy Gradient method with MC roll-out to optimize the policy.\cite{yu2017seqgan} In order to reduce variance in the optimization process, SeqGAN runs the roll-out policy starting from current state till the end of the sequence for multiple times to get the mean return. Here we use an Actor-Critic method with a Critic network instead of MC roll-out to estimate the value of any state, which is computationally more efficient.\cite{bhatnagar2007naturalgradient}
 
The critic network models a state-dependent value $\hat{V}^{G_{\theta}}_{\psi}(S_m)$ for a partially generated sequence $\bm{A}_{1:m}$ under policy $G_{\theta}$. The output of the critic is defined as the expected future return for the current state $S_m = \bm{A}_{1:m}$, which will be given by the discriminator $D$ when a complete sequence $\bm{A}$ is generated.
\begin{equation}
   \hat{V}^{G_{\theta}}_{\psi}(S_m) = \mathbb{E}_{\bm{A}_{m+1:L} \sim G_{\theta}(S_m)} [D_{\phi}(\bm{A})] \label{equ:critic_value}
\end{equation}
The parameters in the critic value function $\hat{V}^{G_{\theta}}_{\psi}(S_m)$ are updated during training by minimizing the mean squared error between the true return $D_{\phi}(\bm{A})$ and the critic value:
\begin{align}
   J(\psi) = \mathbb{E}[(D_{\phi}(\bm{A}) - \hat{V}^{G_{\theta}}_{\psi}(S_m))^2] \label{equ:critic_loss}
\end{align}
The difference between them, $D_{\phi}(\bm{A}) - \hat{V}^{G_{\theta}}_{\psi}(S_m)$, is named the advantage function, which can be used in $G$ training and helps to reduce variance.

The goal of $G$ training is to choose actions based on a policy that maximizes expected return. The object function of $G$ follows Policy Gradient method \cite{sutton2018reinforcement} which can be derived as:
\begin{equation}
   \nabla_{\theta} J(\theta) = \sum_{m=0}^{L-1} \mathbb{E}_{a_m \sim G_{\theta}(a_m | S_m)} [\nabla_{\theta} \log G_{\theta}(a_m | S_m)
   \cdot (D_{\phi}(\bm{A}) - \hat{V}^{G_{\theta}}_{\psi}(S_m))]  \label{equ:policy_gradient_update}
\end{equation}
Because of the independence assumption we made, the policy gradient term can be broken down and written into a categorical cross-entropy and a Gaussian log-likelihood as follows:

\begin{align} 
   & \nabla_{\theta} \log G_{\theta}(a_m | S_m) \nonumber  \\
   =& \nabla_{\theta} [\log Cat(x=x_{m+1}| \alpha_{\theta}(S_{m})) + \log \mathcal{N}(\Delta t = \Delta t_{m+1}| \mu_{\theta}(S_{m}), \sigma^2_{\theta}(S_{m}))] \nonumber \\
  =& \nabla_{\theta} [\mathbb{E}_{x \in \bm{K}} \mathds{1}_{x}(x_{m+1})\Pr(x=x_{m+1}) - \frac{(\Delta t_{m+1}- \mu_{\theta}(S_{m}))^2}{2\sigma^2_{\theta}(S_{m})} - \frac{1}{2} \log(2\pi \sigma^2_{\theta}(S_{m}))]
 \label{equ:log_likelihood}
\end{align}

The goal of $D$ training to use distinguish generated sequences with true sequences from training data. $D_\phi$ is updated through minimizing binary cross-entropy loss. $G$ and $D$ alternatively in GAN training. 

The training data are taken from the positive class $\bm{\Omega}^+$ of our synthetic Ad event dataset $\bm{\Omega}$, which are shown in the section \ref{sec:syn_data}.

Before GAN training, We pre-train $G$ with Maximum Likelihood Estimation (MLE) self-regression on the sequences and pre-train $D$ with binary classification for better convergence. Details about pre-training and GAN training The Pseudo code of the entire process is shown in Algorithm \ref{seqGAN}. 

\begin{algorithm}
\caption{Sequence Generative Adversarial Nets Training with RL}\label{seqGAN}
\textbf{Require:} training dataset $\bm{\Omega}^+$, generator $G_\theta$, discriminator $D_\phi$, critic $\hat{V}^{G_{\theta}}_{\psi}$.
\begin{algorithmic}[1]
    \State Initialize $G_\theta$, $D_\phi$, $\hat{V}^{G_{\theta}}_{\psi}$ with random weights $\theta$, $\phi$, $\psi$
    \State Pre-train $G_\theta$ with MLE self-regression on $\bm{\Omega}^+$.
    \State Generate fake dataset $\bm{\Omega}^{+fake}$ using pre-trained $G_\theta$.
    \State Pre-train $D_\phi$ via minimizing binary cross-entropy on $\bm{\Omega}^+ \cup \bm{\Omega}^{+fake} $
    \Repeat
        \For {$G$-steps}
            \State Generate a batch of fake sequences $\bm{A}^{fake} \sim G_\theta$
            \State Get true rewards $D_{\phi}(\bm{A})$ from discriminator
            \For {$m$ in $1:L$}
                \State $S_m \leftarrow \bm{A}^{fake}_{1:m}$
                \State $a_m \leftarrow (x_{m+1}, \Delta t_{m+1}) \in \bm{A}^{fake}$
                \State $\alpha_{\theta}(S_{m}), \mu_{\theta}(S_{m}), \sigma_{\theta}(S_{m}) \leftarrow G_{\theta}(S_m)$ 
                \State Compute policy gradient as shown in Eq. (\ref{equ:log_likelihood})
                \State Compute value estimate $\hat{V}^{G_{\theta}}_{\psi}(S_m))$ by Eq. (\ref{equ:critic_value})
                \State Compute the advantage $(D_{\phi}(\bm{A}) - \hat{V}^{G_{\theta}}_{\psi}(S_m))$
            \EndFor
            \State Update critic param. $\psi$ by minimizing Eq. \eqref{equ:critic_loss}
            \State Update generator param. $\theta$ via Eq. (\ref{equ:policy_gradient_update})
        \EndFor
        \For {$D$-steps}
            \State Generate a batch of sequences $\bm{A}^{fake} \sim G_\theta$
            \State Sample a batch of sequences $\bm{A}^{true}$ from $\bm{\Omega}^+ $
            \State Train discriminator $D_\phi$ on $\bm{A}_{fake} \cup \bm{A}_{true}$ and update param. $\phi$ via minimizing binary cross-entropy
        \EndFor
    \Until terminate condition satisfied
\end{algorithmic}
\end{algorithm}

\subsection{An Approximation with Gumbel-Softmax Distribution}
Beside RL, we also tried to overcome the gradient updates problem for discrete token in GAN using Gumbel-Softmax reparametrization.
We use the same generator $G$ and discriminator $D$ setups as described in section \ref{sec:rl_train}, except that the generator policy $G_{\theta}(a_m|S_m)$ is different from that in (\ref{equ:hybrid_policy}). 
For the continuous part, we no longer sample time intervals from a parametrized Normal distribution, but directly take $G$ outputs as the next time interval.
\begin{equation}
   a^{\Delta t}_m = \Delta t_{m+1} = \Delta t_{\theta}(S_m)
\end{equation}
For the discrete part, in the forward pass of training the generator $G$, we add a Gumbel noise to the probability distribution of event types at each step $m$, and use argmax operator to sample the next event type $x_{m+1}$:
\begin{equation}
    a^{x}_m = x_{m+1} = \argmax_{i}(\log (\alpha_{\theta}(S_{m})_i) + g_i) \quad \text{for} \quad i = 1,\dots,|\bm{K}|
     \label{equ:gumbel_policy_forward}
\end{equation}
where $\tau$ is the temperature and $g$ is a random variable with a standard Gumbel distribution:
\begin{equation}
g = -\log(-\log(U)), \quad \text{where} \quad U \sim \text{Uniform}([0,1])
\end{equation}
\noindent
In the backward pass of $G$ training, we reparametrize the categorical distribution using a Gumbel random variable $g$ to create a differentiable approximation of the discrete representation of $a^{x}_m$ to calculate gradients:
\begin{align}
    \text{Pr}(a^{x}_m =x_i, x_i \in \bm{K}| S_{m}) = &\frac{\exp{((\log (\alpha_{\theta}(S_{m})_i) + g_i) / \tau)}}{\sum_{j=1}^{k} \exp{((\log (\alpha_{\theta}(S_{m})_j) + g_j) / \tau)}}   \\
    & \quad \text{for} \quad i = 1,\dots,|\bm{K}| \nonumber \label{equ:gumbel_policy_backward}
\end{align}
\noindent

After the Gumbel-Softmax reparametrization, we can train the multi-type GAN with discrete event types using a similar approach in \cite{kusner2016gans}.

\section{Data Experiments}
Due to data privacy laws (e.g. GDPR \footnote{General Data Protection Regulation}, CCPA \footnote{California Consumer Privacy Act}), and to protect confidential details of the Unity Ads Exchange and Fraud Detection service, we opt not to use real-world ad events data in this study to avoid releasing user behavior patterns to the public. While anonymizing the real-world dataset can hide users' identities, it cannot disguise the users' behavior patterns and distributions. Fraudsters can easily employ bots to simulate the features of real users to bypass fraud detection systems, if given access to the real data.

Instead, we conduct our experiments on a synthetic dataset, which contains simplified data patterns we observed and abstracted from real-world ad events. The design philosophy is explained in Section~\ref{sec:syn_data}. The synthetic dataset and code used to generate it are publicly available\footnote{\url{https://github.com/project-basileus/multitype-sequence-generation-by-tlstm-gan}}. 

\subsection{Synthetic Dataset}\label{sec:syn_data}
We define the synthetic dataset as $\bm{\Omega}$. There are $4$ types of hypothetical ad events in $\bm{\Omega}$, shown as $\bm{K} = \{ a, b, c, d\}$. Each sequence in the synthetic dataset $\bm{\Omega}$ has a uniform length $L=20$. A step at $m$ corresponds to a tuple of event type and time interval, $(x_m, \Delta t_m)$, where $x_m$ is sampled uniformly from $\bm{K}$, and $\Delta t_m$ is sampled from a Chi-Square distribution with the degree of freedom conditioned on $x_m$, i.e.:
\begin{equation}
x_m \sim \text{Uniform}\{a,b,c,d\} \quad
\Delta t_m \sim \mathcal{X}^2(k), \quad k = \left\{
        \begin{array}{ll}
            10 & \quad \text{if} \quad x_m = a \\
            20 & \quad \text{if} \quad x_m = b \\
            40 & \quad \text{if} \quad x_m = c \\
            80 & \quad \text{if} \quad x_m = d \\
        \end{array}
    \right.
\end{equation}

One example of a complete synthetic sequence is as below:

\begin{align*}
\bm{A}_{e.g.} = [& (a, 5), (a, 22), (b, 27), (c, 44), (c, 43), \\
 & (d, 87), (b, 30), (c, 36), (d, 75), (c, 28), \\
 & (a, 9), (b, 24), (a, 9), (c, 40), (b, 29), \\
 & (c, 37), (a, 10), (b, 19), (c, 26), (b, 7)]
\end{align*}

There are two classes in $\bm{\Omega}$, the positive class $\bm{\Omega}^+$ and the negative class $\bm{\Omega}^-$. As the two classes can be highly imbalanced in real-world Ad events data (e.g. fraud/non-fraud, buyer/non-Buyer, conversion/non-conversion, etc.), the positive class is the minority in $\bm{\Omega}$, with a positive-to-negative ratio of $1:500$. A positive sequence has the following properties:
\begin{enumerate}
  \item The time delay between any two consecutive events of the same event type is greater than or equal to $20$. 
  \item Each $d$ event is paired with one and only one previous $c$ event. Each $c$ event can be paired with at most one $d$ event after it.
  \item The time delay between any two paired $c$ and $d$ events is smaller than or equal to $200$.
\end{enumerate}
Sequences failing to have all 3 properties above are considered negative. The positive class $\bm{\Omega}^+$ is the training dataset. We train a GAN to generate data points from the minority class with the above properties. We will employ them as an oracle to evaluate the quality of GAN-generated sequences, as described in section \ref{sec:eval_matric}. 

The design philosophy of the synthetic dataset is to simulate real-world patterns with as much fidelity as possible while hiding real parameters to prevent reverse-engineering by fraudsters. Specifically, the hypothetical ad events $\{ a, b, c, d\}$ mimic four typical real ad events: starts, views, clicks, and installs. Real-world time delay between ad events follows a long-tail distribution, while in the synthetic dataset, it is modeled with a Chi-Square distribution conditioned on the preceding event type. Moreover, the three properties of a positive sequence are also abstracted from real-world data patterns: property 1 detects high-frequency attacks; property 2 describes the ad attribution process between clicks and installs; property 3 checks the validity of an attribution window. Ad attribution refers to the process of determining the user actions that led to the desired outcome between the click of the ad and the conversion.

\subsection{Evaluation Metric} \label{sec:eval_matric}
In the last few years, several different evaluation metrics for GANs have been introduced in the literature. Among them, Fréchet Inception Distance (FID) \cite{heusel2017gans} has been used extensively \cite{devries2019evaluation}.
However, this only captures the numerical part of a sequence, but our sequences are multi-type containing both the discrete categorical part (event type) and the continuous numerical part (time interval). Thus, we propose using multiple metrics to measure the quality of generated sequences. We use Mean Absolute Deviation (MAD) to measure the discrete event types, and use FID to evaluate the continuous time intervals. In addition, we employ an oracle score based on the known properties in the training data to measure the similarity between generated sequences and the training data. The arrows ($\uparrow\downarrow$) show the improvement directions.


\textbf{MAD $\downarrow$}. We propose using MAD to evaluate the statistical dispersion between the categorical part (i.e., the event types) of the generated multi-type sequences and that of the training data. We use the training dataset $\bm{\Omega}^+$ as the comparison base, and then one-hot encode the event types of training sequences to calculate the medians at each step $m$. Median is known to be more robust to noise and fits our need to have categorical values as opposed to mean.

The MAD score of any batches of generated sequences $\bm{B}$ is computed as the mean absolute deviation of each sequence from the base medians, shown as below as  MAD can be computed using:

\begin{equation}
  MAD(\bm{B}) = \frac{1}{|\bm{B}|}\sum_{\bm{A} \in \bm{B}}\sum_{m=1}^{L}\left|x_m^{\bm{A}} - \Tilde{E}_{m}(\bm{\Omega}^+)\right| 
\end{equation}
where $\bm{B}$ is a batch of generated sequences, $|\bm{B}|$ is the batch size, $\bm{A}$ is a sequence of length $L$ in $\bm{B}$, $x_m^{\bm{A}}$ is the event type of step $m$ in $\bm{A}$, $\Tilde{E}_{m}(\bm{\Omega}^+)$ is the base median of the event types at step $m$ across the training dataset $\bm{\Omega}^+$.

\textbf{FID $\downarrow$}. Similarly to MAD, we use FID to measure the distance between the numerical part (i.e., the time intervals) of the multi-type sequences and that of the training data. This score focuses on capturing certain desirable properties including the quality and diversity of the generated sequences. FID performs well in terms of robustness and computational efficiency \cite{borji2019pros}. The Fréchet distance between two Gaussians is defined as:
\begin{equation} \label{eq:fid}
  FID(x,g) =  \norm{\mu_x - \mu_g}^2_2 + Tr \left(\Sigma_x + \Sigma_g - 2 \left( \Sigma_x\Sigma_g \right)^{\frac{1}{2}} \right)
 \end{equation}
where $\left(\mu_x, \Sigma_x \right)$ and $\left(\mu_g, \Sigma_g \right)$ are the means and covariances for the training and generated data distribution, respectively.

\textbf{Oracle $\uparrow$}. One of the most direct ways to measure the quality of a generated sequence is to check whether it has the known data properties of the positive class (described in section \ref{sec:syn_data}). For a batch of generated sequences, we calculate the percentage of sequences having all 3 properties of the positive class over all sequences, and then use this ratio as the oracle score. For example, for a data batch from the training dataset $\bm{\Omega}^+$, the oracle score is $1$. The oracle score is a metric taking both the continuous and discrete part of a sequence into consideration.

\subsection{Experiment Setup}
We take $4000$ samples from the $\bm{\Omega}^+$ dataset defined in section \ref{sec:syn_data} for model training. As is described in Algorithm \ref{seqGAN}, we first pre-train $G$ and $D$ and then start GAN training from the pre-trained $G$ and $D$. 
We define the following terms to describe the generator at different training phases:
\begin{itemize}
\item $G0$: Generator with initial random model parameters.
\item $G1$: Generator pre-trained using MLE self-regression.
\item $G2$: Generator after GAN training.
\end{itemize}
The ratio between $G$ training steps and $D$ training steps is set to $1:1$. Both $G$ and $D$ have the same batch size $256$, and use the Adam optimizer with learning rate $10^{-4}$. 

During the pre-training and training processes, we evaluated the performance of the trained generator $G$ after some steps. The trained generator was then used to generate a batch of data points and the batch evaluated according to the metrics defined in section \ref{sec:eval_matric}.

To avoid mode collapse and convergence problems, we used several techniques including label smoothing and noisy labels \cite{salimans2016improved} in GAN training. In RL training, we added entropy regularizers \cite{dieng2019prescribed} to the reward for discrete token and continuous time interval generation to avoid over-fitting.

\subsection{Experiment Results}
\begin{figure}[ht]
\centering
\subfigure{\includegraphics[width=.34\textwidth]{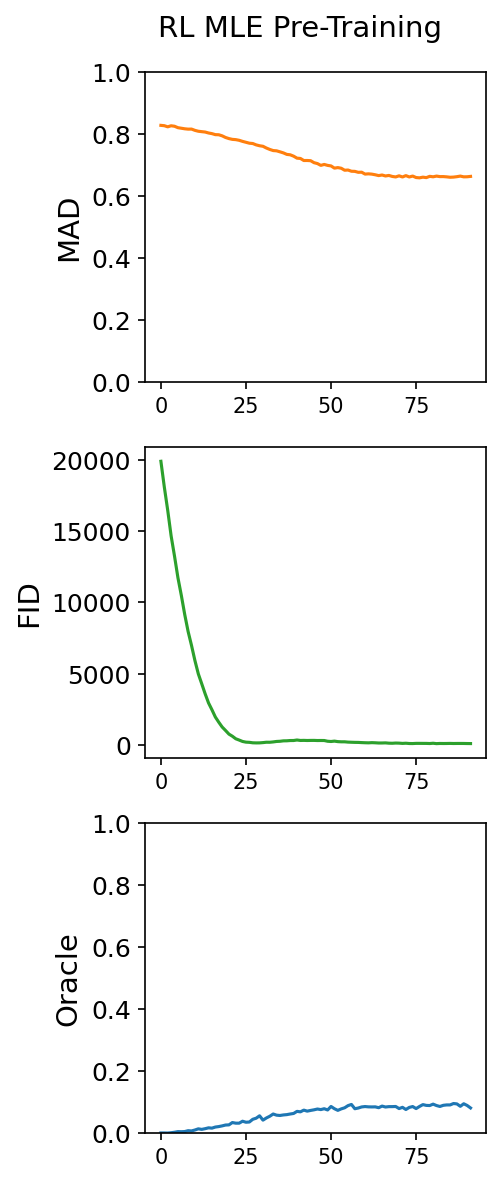}}
\subfigure{\includegraphics[width=.34\textwidth]{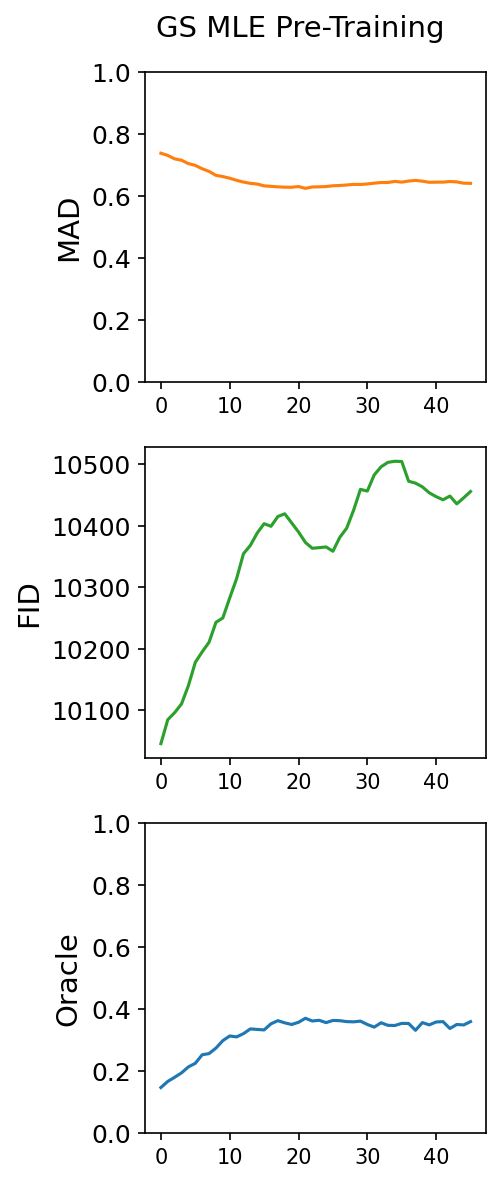}}
\caption{Metrics of generated sequences over pre-training steps for Reinforcement Learning (RL) and Gumbel-Softmax (GS).}
\label{fig:pretrain}
\end{figure}

\begin{figure}[ht]
\centering
\subfigure{\includegraphics[width=.33\textwidth]{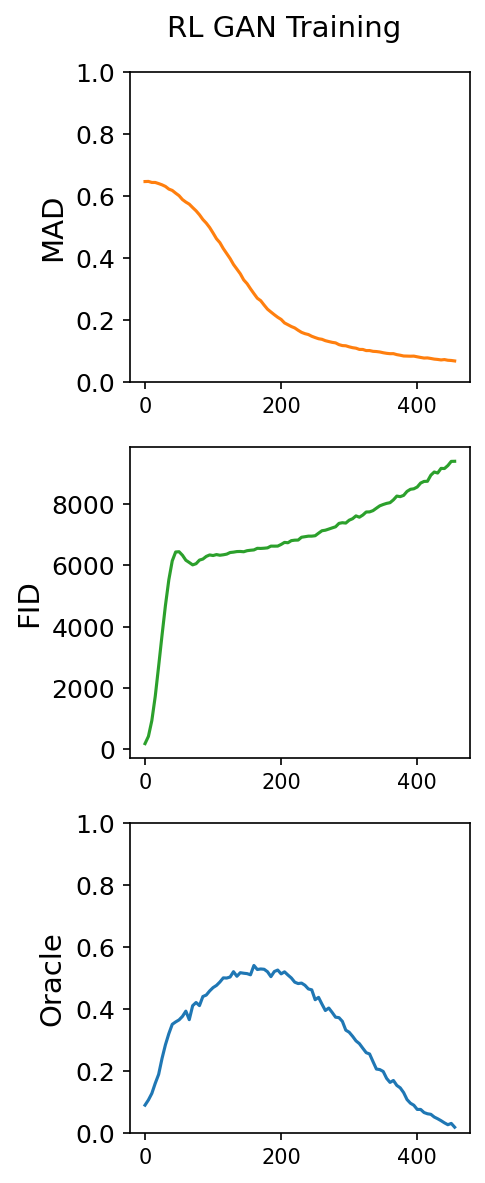}}
\subfigure{\includegraphics[width=.34\textwidth]{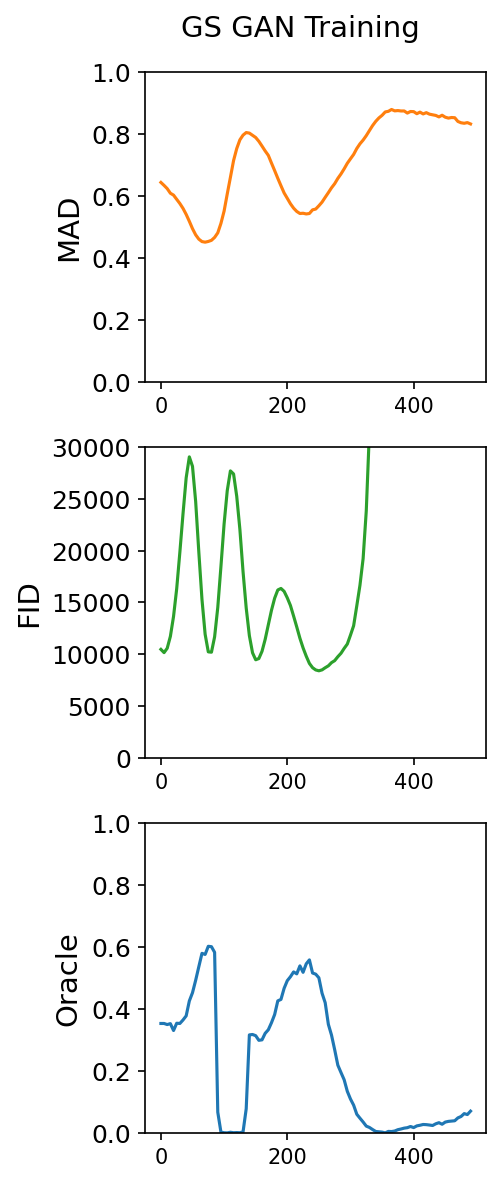}}
\caption{Metrics of generated sequences over training steps for Reinforcement Learning (RL) and Gumbel-Softmax (GS).}
\label{fig:train}
\end{figure}

Table \ref{tab:metric_pos} shows the evaluation metric values of the sequences generated by $G$ at different phases of training. The MAD, FID score are calculated respectively using data sampled from $\bm{\Omega}^+$ as the base for the comparisons. 
\begin{table}[tb]
\centering
  \caption{Oracle metrics calculated using $\bm{\Omega^{+}}$ as base}
  \label{tab:metric_pos}
  \begin{tabular}{|c||c|c|c||c|c|c|}
    \hline
      & \multicolumn{3}{|c||}{Reinforcement Learning (RL)} & \multicolumn{3}{|c|}{Gumbel-Softmax (GS)}\\
    \hline
    \textbf{Samp.}  & \textbf{MAD $\downarrow$} & \textbf{FID $\downarrow$} &\textbf{Oracle $\uparrow$}& \textbf{MAD $\downarrow$} & \textbf{FID $\downarrow$} & \textbf{Oracle $\uparrow$} \\
    \hline
    \textbf{G0}  & 0.8265 & 19892.4782 & 0.0015    & 0.7368 & 10045.2759 & 0.1477\\
    \textbf{G1}  & 0.6622 & \textbf{101.7972} & 0.0820   & 0.6399 & 10455.6409 & 0.3600    \\
    \textbf{G2}  & \textbf{0.2849} & 6495.2955 & \textbf{0.5407}
          & \textbf{0.5427} & \textbf{9111.5298} & \textbf{0.55875}
   \\
  \hline
\end{tabular}
\end{table}

The curves of evaluation metrics during pre-training and training are shown in Figure \ref{fig:pretrain} and Figure \ref{fig:train}, respectively.

The results in table \ref{tab:metric_pos} demonstrate that the sequences generated by GAN-trained $G2$ have a significantly higher oracle score than that generated by the MLE pre-trained generator $G1$ and randomly initialized generator $G0$, for both RL and Gumbel-Softmax training. This indicates that the generator is able to learn the intrinsic patterns and properties in the training data $\bm{\Omega^{+}}$, and is able to mimic these patterns to deceive the discriminator. 

From the perspective of metric curves, we noticed that in the pre-training of RL, the FID score of the generator decayed sharply from around $20,000$ to around $100$, while the the improvements of MAD score and oracle score were stalling. It suggested that the MLE training was over-fitting in learning the continuous distribution of the time interval $\Delta t$, while paying much less effort to learn the patterns in the discrete event type $x$, and the relationships and hidden connections between the continuous and the discrete parts.

Comparing the performance of RL and Gumbel-Softmax training approaches, we found that the RL approach converged faster in pre-training and training with smoother metrics curves, but it was vulnerable to over-fitting and Gaussian model collapsing. Meanwhile, the Gumbel-Softmax approach converged slower with more curve oscillations, but it was less prone to over-fitting, even with the entropy regularizers in reward. 

\section{Conclusions} 
In this paper, we have described, trained, and evaluated a novel methodology for generating artificial sequences with multi-type tokens. As this task poses new challenges, we have presented and compared the policy gradient (RL) and Gumbel-Softmax approaches for training a multi-type GAN. The generator proposed in this paper is capable of generating multi-type temporal sequences with non-uniform time intervals. We have also proposed using multiple criteria to measure the quality of the generated sequences. Experiments demonstrate that the generated multi-type sequences contain the desired properties.

Furthermore, we compared the performance of our generator for both RL and GS approaches with data from our carefully designed synthetic dataset. We concluded that the SeqGAN-trained generator has a higher performance compared to pre-trained generators using self-regression MLE, measured by multiple criteria including MAD, FID, oracle scores that are appropriate for evaluating multi-type sequences.

\section*{Acknowledgments}
The authors would like to thank Unity for giving the opportunity to work on this project during Unity's HackWeek 2020.
%
%

\bibliographystyle{splncs04}
\bibliography{main}

\end{document}